\documentclass[11pt]{article}

\usepackage[final]{acl}

\usepackage{times}
\usepackage{latexsym}
\usepackage{multirow}
\usepackage{booktabs}
\usepackage{graphicx}
\usepackage{svg}
\usepackage{amsmath,amssymb,amsfonts}
\usepackage{adjustbox}
\usepackage{subcaption}
\usepackage{comment}

\usepackage[T1]{fontenc}

\usepackage[utf8]{inputenc}

\usepackage{microtype}

\usepackage{inconsolata}

\usepackage{graphicx}

\title{Orthogonal Representation Editing: Decoupling Semantic Entanglement \\in Batch Knowledge Editing of LLMs}

\author{
    \textbf{Wenhao Yu\textsuperscript{1,2}},
    \textbf{Zhicong Lu\textsuperscript{3}},
    \textbf{Bo Lv\textsuperscript{4}},
    \textbf{Fangyin Ma\textsuperscript{1}},
    \textbf{Kaiwen Wei\textsuperscript{5}},
    \textbf{Shihao Yang\textsuperscript{1}},
    \textbf{Nayu Liu\textsuperscript{1$\dagger$}}
    \\
    \\
     \textsuperscript{1}School of Computer Science and Technology, Tianjin University
    \\
     \textsuperscript{2}Kexin Technology, 
     \textsuperscript{3}University of Chinese Academy of Sciences
    \\
     \textsuperscript{4}Tencent Hunyuan, \textsuperscript{5}College of Computer Science, Chongqing University
    \\
    \texttt{
        nyliu@tju.edu.cn
     }
}

\begin{document}
\maketitle
\renewcommand{\thefootnote}{}
\footnote{$\dagger$ Corresponding author.}
\begin{abstract}
Knowledge editing aims to efficiently update factual information in Large Language Models (LLMs) without full retraining. However, existing methods still suffer from performance degradation in batch knowledge editing. We identify that semantic representation entanglement, such as overlapping concepts and shared syntactic patterns, accumulates interference in the representation space and reduces editing precision.
To bridge this gap, in this paper, we propose Orthogonal Representation Editing (ORE), which performs edits in the hidden representation space of LLMs by constructing a general semantic subspace and enforcing orthogonal constraints on edit vectors, effectively decoupling semantic entanglement. Furthermore, we introduce a gated non-linear representation head to enable adaptive learning of editing locations and precise control over knowledge injection. Extensive experiments show that ORE outperforms existing methods and achieves superior performance in cross-lingual knowledge editing scenarios. We release our code at \url{https://github.com/YVVH/ORE}.
\end{abstract}

\section{Introduction}

Large Language Models (LLMs) have demonstrated strong capabilities in question answering and reasoning \cite{mann2020language, brown2020language}. Despite these advances, their parameters remain inherently static, limiting their ability to accommodate the continual evolution of real-world knowledge. Knowledge editing has therefore emerged as a crucial research direction for efficiently updating specific facts in pretrained models without retraining from scratch \cite{yao2023editing, wang2024knowledge, gupta2024model}.

\begin{figure}[h]
  \centering
  \includegraphics[width=\linewidth]{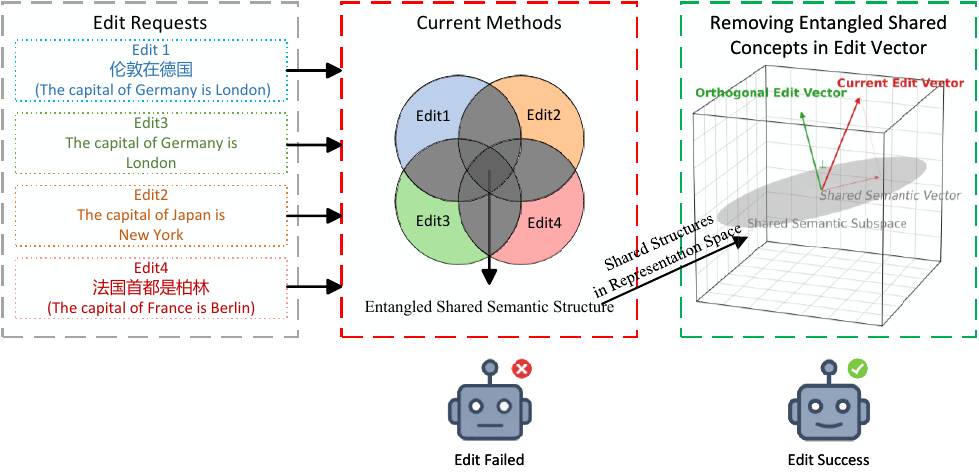}
  \caption{Edits with different factual targets activate overlapping regions in representation space, finally leading to interference.}
  \label{fig:overview}
\end{figure}

Existing knowledge editing methods can be broadly categorized into two paradigms. Parameter-preserving methods \cite{huang2023transformer, REMEDI, GRACE, SAKE, BaFT, deck} maintain a frozen backbone and introduce auxiliary modules to update knowledge, while parameter-modifying methods \cite{ROME, AlphaEdit, jiang2025anyedit} directly locate and update a subset of model parameters to inject new facts. Notably, MEMIT \citep{MEMIT} extends parameter-modifying approaches to batch knowledge editing, enabling the simultaneous injection of thousands of knowledge entries by distributing update residuals across multiple layers.

Despite their empirical success, existing knowledge editing methods exhibit performance degradation in such batch editing settings \cite{MEMIT}. To explore the reasons, we observe that edits targeting different facts are not isolated in the representation space but instead occupy overlapping regions. For example, as shown in Fig.~\ref{fig:overview}, the edits “The capital of Japan is Tokyo” and “The capital of France is Berlin” follow the same template “[Country’s capital] is [City]” and invoke the shared concept of “capital,” leading to interference within a common region of the representation space.

We attribute this degradation to \textbf{semantic representation entanglement}. In practice, many edit requests share overlapping concepts or general syntactic templates, which induces interference among their corresponding representations in the hidden space. Existing methods struggle to disentangle fact-specific information from such general semantic structures; as the editing size grows, accumulated interference substantially degrades editing precision. This issue is further exacerbated in cross-lingual settings \cite{beniwal2024cross, LangEdit}, where shared multilingual semantic spaces allow edits to propagate along semantic directions into non-target languages, resulting in unintended cross-lingual interference. Empirical evidence supporting these observations is presented in \textbf{Section~\ref{sec:observation}}.

Motivated by these observations, we propose \textbf{Orthogonal Representation Editing (ORE)}, a representation-based knowledge editing framework guided by geometric constraints \cite{REMEDI, cai2024edit, KDE}. ORE operates directly in the hidden representation space and aims to decouple editing directions, thereby mitigating interference induced by shared semantic patterns and enabling more reliable knowledge updates \cite{wang2025decoupling}.
Specifically, ORE leverages a set of irrelevant but structurally similar samples to estimate a \emph{general semantic subspace}. Each edit vector is then orthogonalized by subtracting its projection onto this subspace, ensuring that knowledge updates occur along directions independent of shared semantics. 
To realize orthogonal editing in practice, ORE builds upon representation fine-tuning (ReFT)~\cite{ReFT}. In addition, to move beyond linear interventions and manual positional priors, ORE introduces a gated non-linear representation head that adaptively determines \emph{when} and \emph{where} to intervene, enabling precise knowledge injection with minimal impact on general capabilities.
Extensive experiments demonstrate that ORE consistently outperforms existing methods and remains robust in cross-lingual knowledge editing scenarios.
In summary, the main contributions of this paper are as follows:

\begin{itemize}

    \item We identify \textbf{semantic representation entanglement} as a fundamental limitation of batch knowledge editing, where interference accumulates in the representation space and degrades editing performance.

    \item We propose Orthogonal Representation Editing (ORE), which constructs a general semantic subspace and performs orthogonal, gated interventions in the representation space to decouple shared semantic entanglement.

    \item Extensive experiments demonstrate that ORE achieves strong and consistent performance across multiple benchmarks and remains effective in challenging cross-lingual knowledge editing scenarios.

\end{itemize}

\section{Observation of General Semantic Representation Entanglement}
\label{sec:observation}
To empirically verify the hypothesis that general semantic structure entanglement leads to performance degradation in batch editing, we designed a controlled experiment to compare the performance differences of existing methods on random data versus data with high general semantic entanglement.

\subsection{Data Settings}

\noindent\textbf{Entangled Samples:} We utilized Gemini 3 to construct 200 cross-lingual samples with identical syntactic structures and highly correlated semantics based on the theme of "Capital," including 100 French samples and 100 Chinese samples. All samples follow the pattern "The capital of [Country] is [City]." Under this setting, different subject entities activate a general semantic structure within the model's representation space. The entangled samples are divided into three groups: the French group (100 French samples), the Chinese group (100 Chinese samples), and the Cross-lingual group (50 French and 50 Chinese samples).

\noindent\textbf{Random Samples:} We randomly sampled 100 entries from the ZsRE \cite{zsre} dataset and translated them into French and Chinese languages. These samples cover diverse relation types and distinct semantic categories, possess strong semantic independence, and serve as a control group.

\subsection{Observations}

\begin{figure}[t]
  \centering
  \includegraphics[width=0.5\textwidth]{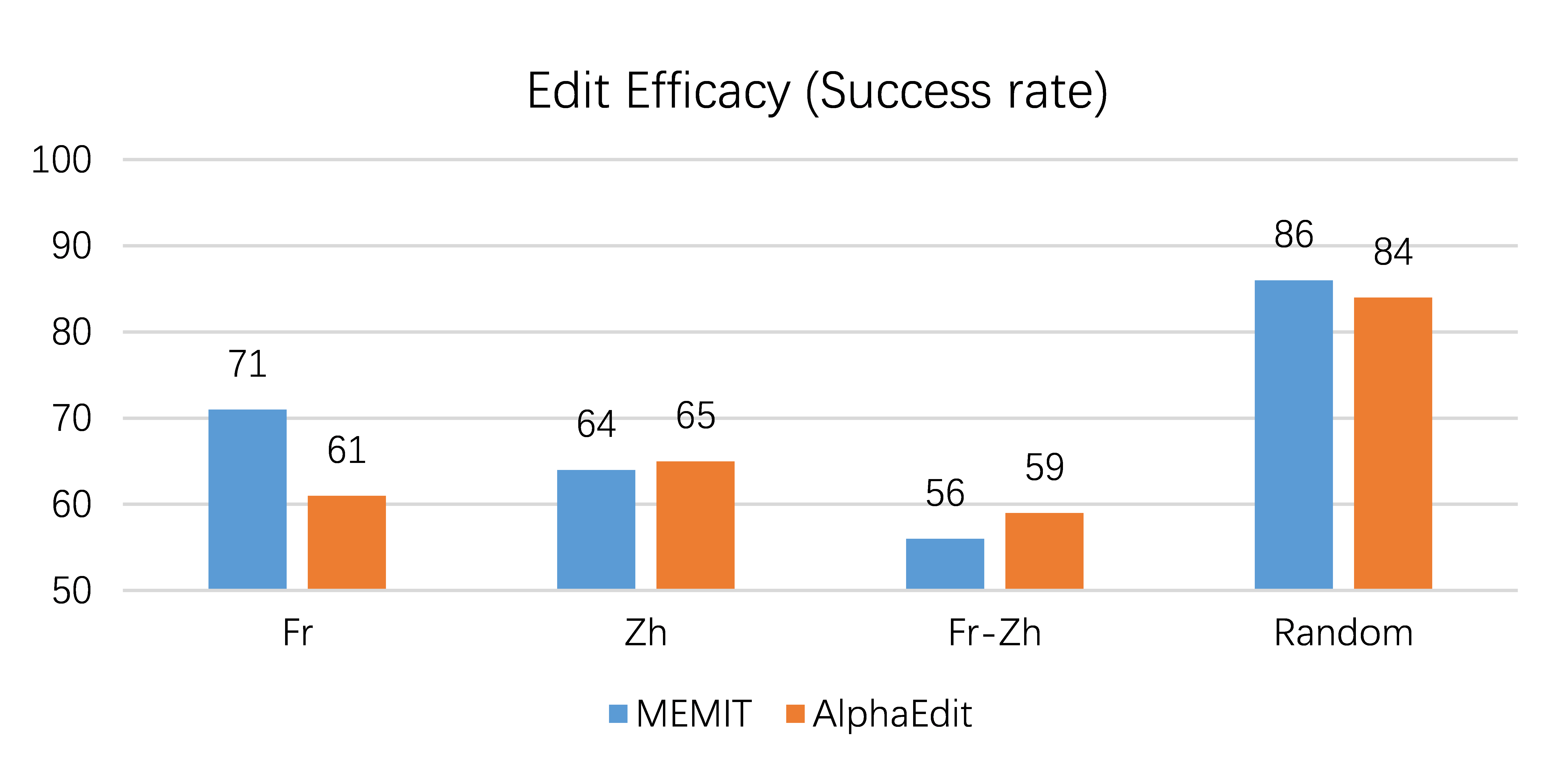}
  \caption{Editing efficacy of MEMIT and AlphaEdit on entangled (Fr, Zh, Fr-Zh) and random samples. Performance drops on entangled samples, especially in cross-lingual (Fr-Zh) settings.}
  \label{fig:observation}
\end{figure}

We employed MEMIT \cite{MEMIT} and AlphaEdit \cite{AlphaEdit}, currently representative editing methods as baselines to edit the aforementioned two groups of data on the LLaMA-3-8B model and recorded the Editing Success. As shown in Figure \ref{fig:observation}, MEMIT demonstrated robust performance on the random group, achieving a high success rate of 86\%. However, on the entangled samples, MEMIT's editing accuracy declined sharply: the success rate for the French group was only 71\%, for the Chinese group it was 64\%, and for the Cross-lingual group, it further dropped to 56\%. Similar observations also occurred in AlphaEdit, which indicate that performance degrades when facing samples with general concept entanglement.

This result supports our hypothesis: in batch editing, shared general semantic structures cause update vectors to conflict and accumulate noise within the representation subspace, thereby leading to a decline in the performance of existing methods.

\section{Methodology}

\begin{figure*}[t]
  \centering
  \includegraphics[width=0.9\textwidth]{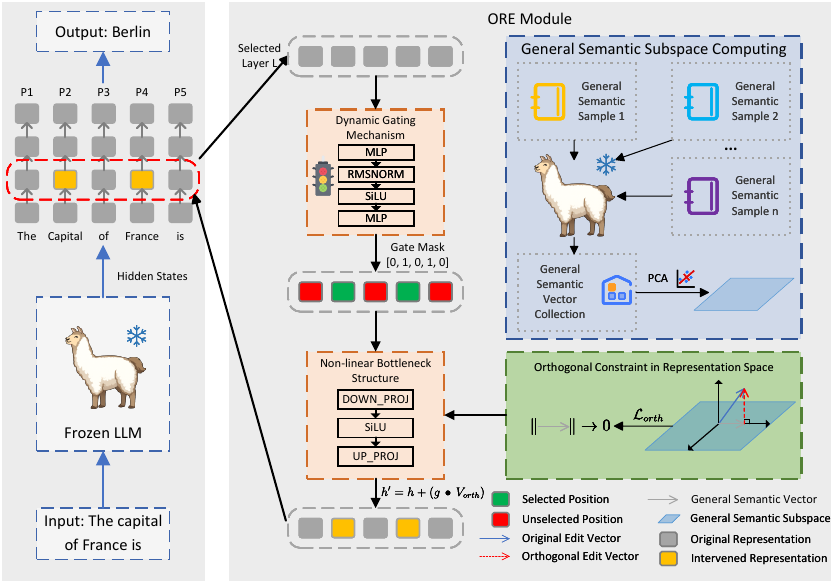}
  \caption{Overview of the proposed ORE framework. It edits a frozen LLM by applying gated, non-linear representation interventions at selected layers, orthogonalizing edit vectors against a general semantic subspace, and updating hidden states with the resulting orthogonalized edits.}
  \label{fig:method}
  \vspace{-6pt}
\end{figure*}

\subsection{Overview of ORE}
Motivated by the empirical observations in Section \ref{sec:observation}, we propose ORE, a representation editing framework based on geometric constraints. As shown in Figure \ref{fig:method}, ORE comprises: (1) Representation Subspace Orthogonalization, which aims to explicitly construct a general semantic subspace and strip away general semantic noise from edit vectors via orthogonal projection, thereby mitigating the entanglement between knowledge items from a geometric perspective. (2) Gated Non-linear Representation Head, which is designed to adapt representation fine-tuning for vector editing. It utilizes a non-linear bottleneck and a dynamic gating mechanism to perform fine-grained intervention and adaptive injection on representation vectors.

\subsection{Representation Subspace Orthogonalization}
To address general semantic entanglement in batch knowledge editing, we explicitly model shared semantics as a subspace in the representation space and constrain edits to lie outside this subspace.

\subsubsection{Construction of General Semantic Subspace}
To extract the general semantic structure involved in batch editing, we construct a sample set $D_{gs}$ consisting of $N$ samples that share similar structures with the target edits while being factually unrelated. Concretely, these samples are randomly drawn from the ZsRE \cite{zsre} and CounterFact \cite{ROME} datasets, excluding all instances used for training or evaluation, following the same prompt pattern as the target edits, but involving different subject–object pairs, ensuring that no target facts are included.

We extract their hidden states at the target layer $l$ on the backbone model, denoted as $H_{gs}=\{h^{(l)}_1, h^{(l)}_2, ..., h^{(l)}_N\}$. To capture the dominant directions corresponding to these shared semantics, we perform Principal Component Analysis (PCA) \cite{abdi2010principal} on $H_{gs}$ and select the top $k$ principal components to form an orthogonal basis matrix $U_{gs} \in \mathbb{R}^{d \times k}$. The subspace spanned by $U_{gs}$ is defined as the General Semantic Subspace $S_{gs}$.

\subsubsection{Orthogonal Constraint on Edit Vectors}
To operationalize subspace orthogonalization, we compute two representations of the source input: the source representation produced by the original frozen model, denoted as $h^{(l)}_{src,orig}$, and the source representation produced by the current edited model, denoted as $h^{(l)}_{src}$. Moreover, $h^{(l)}_{pred}$ and $h^{(l)}_{alt}$ denote the representations at layer $l$ corresponding to the model’s prediction statement and the desired alternative statement, respectively. Specifically, these representations are obtained by using the declarative statement of the target knowledge as the input prompt and extracting the hidden states at the last token position. For instance, given an edit request where the prompt is "What university did Watts Humphrey take part in?", if the model originally predicts "Trinity College" but the target is "University of Michigan", we construct the declarative statements "Watts Humphrey attended Trinity College." for $h^{(l)}_{pred}$ and "Watts Humphrey attended the University of Michigan." for $h^{(l)}_{alt}$. We then feed these full sentences into the model and extract the vectors from their respective final token positions \cite{ROME}.
The edit vector $\Delta^{(l)}$ is computed as:

\begin{equation}
    \Delta^{(l)} = h^{(l)}_{alt} - h^{(l)}_{pred},
\end{equation}
which captures the direction required to modify the model’s behavior from the current prediction toward the target fact.
We then remove the projection of $\Delta^{(l)}$ onto the general semantic subspace $S_{gs}$ to obtain an orthogonalized edit direction:
\begin{equation}
    \Delta^{(l)}_{orth} = \Delta^{(l)} - U_{gs} U_{gs}^{\top} \Delta^{(l)},
\end{equation}
which is subsequently used to define the target representation as
\begin{equation}
    h^{(l)}_{tgt} = h^{(l)}_{src,orig} + \Delta^{(l)}_{orth}.
\end{equation}
The training objective then encourages $h^{(l)}_{src}$ to align with $h^{(l)}_{tgt}$.


\subsection{Non-linear Gated Representation Fine-tuning}
To enable vector-level intervention within the model and implement Representation Subspace Orthogonalization, we utilize ReFT \cite{ReFT} to execute this process. ReFT in its standard form was not originally designed for knowledge editing, and thus provides limited support for fine-grained and adaptive intervention. Therefore, ORE introduces a non-linear bottleneck structure to finely adjust semantic expressiveness and incorporates a dynamic gating mechanism to enable adaptive selection of intervention positions.

\subsubsection{Non-linear Bottleneck Structure}
To enhance the ability of edit vectors to capture fine-grained semantic features, we designed a non-linear bottleneck structure based on low-rank projection \cite{hu2022lora}.
Specifically, for the input hidden state $h^{(l)}$, it is first mapped to a low-dimensional manifold via a bias-free dimensionality reduction matrix $W_{down} \in \mathbb{R}^{r \times d}$, then passed through a SiLU non-linear activation function, and finally mapped back to the original representation space via a bias-free dimensionality expansion matrix $W_{up} \in \mathbb{R}^{d \times r}$, formalized as follows:
\begin{equation}
v_{edit} = W_{up} \cdot \text{SiLU}(W_{down} \cdot h^{(l)}),
\end{equation}
where $d$ is the hidden layer dimension, $r$ is the bottleneck rank, and $r \ll d$. 

\subsubsection{Dynamic Gating Mechanism}
To overcome the flexibility limitations imposed by manually specifying edit positions and to minimize interference with irrelevant knowledge, we introduce a dynamic gating network parallel to the bottleneck structure. This module aims to adaptively generate a binary gating coefficient $g \in \{0, 1\}$ based on the current hidden state, thereby achieving precise and sparse intervention at specific knowledge positions.
Specifically, the input state $h^{(l)}$ is first projected to an intermediate dimension via a linear layer, followed by normalization, and then passed through a SiLU activation function to obtain gating features. A final linear projection followed by a sigmoid function $\sigma$ then produces a soft gating score $s$ in the range of $(0, 1)$. The calculation process is defined as follows:
\begin{equation}
s = \sigma(W_{g2} \cdot \text{SiLU}(\text{Norm}(W_{g1}h^{(l)}))),
\end{equation}

To obtain the final discrete gating coefficient $g$, we introduce an indicator function $I(\cdot)$ to apply threshold truncation to $s$. Setting the threshold as $\tau$, $g$ is defined as:
\begin{equation}
g = I(s > \tau) = \begin{cases} 1, & \text{if } s > \tau \\ 0, & \text{if } s \leq \tau \end{cases}.
\end{equation}

We define the intervention function $\Phi(h^{(l)})$ as:

\begin{equation}
\Phi(h^{(l)}) = h^{(l)} + g \cdot v_{edit}.
\end{equation}

To ensure gradient continuity through the non-differentiable indicator $I(\cdot)$, we adopt the straight-through estimator (STE) during training.
\subsection{Loss Functions}
ORE comprises the following supervision mechanisms to facilitate representation fine-tuning: 

\textbf{Orthogonal Projection Loss:} To constrain the model's output vectors to be orthogonal to the general semantic subspace, we treat the sum of the original representation and the orthogonalized edit vector as the target direction. We use cosine similarity to constrain the current model's output representation $h_{src}^{(l)}$ to approach this target:
\begin{equation}
\mathcal{L}_{orth} = 1 - \cos(h_{src}^{(l)}, h_{src,orig}^{(l)} + \Delta^{(l)}_{orth}).
\end{equation}

\textbf{Gating Supervision Loss:} To ensure that the dynamic gating network can precisely localize the key positions where knowledge is stored while remaining silent in non-key regions and on irrelevant samples, we employ Binary Cross-Entropy (BCE) loss for explicit supervision. 
Concretely, we utilize the syntactic structure information of the input text to construct a target gating mask $m \in \{0, 1\}^T$, where $T$ is the sequence length. The supervision signals are divided into two scenarios: First, we set the mask values corresponding to the subject \cite{ROME, MEMIT} token positions to 1 and the remaining positions to 0. This guides $s$ to approach 1 in the subject region, thereby activating the indicator function to output $g=1$. For non-subject regions in edit samples, as well as all irrelevant samples used to preserve general capabilities, we set the mask values entirely to 0. This guides the gating network to output low scores in these regions, thereby closing the intervention channel.

Based on the above definitions, the gating loss $\mathcal{L}_{gate}$ is defined as the BCE loss between the predicted soft gating score $s$ and the target mask $m$:
\begin{equation}
\mathcal{L}_{gate} = -\frac{1}{T}\sum_{t=1}^{T}[m_t\log(s_t)+(1-m_t)\log(1-s_t)].
\end{equation}
where $s_t$ is the soft gating score.

In addition, cross-entropy and KL-divergence losses are employed to enforce the desired output behavior while preserving locality. Specifically, the KL-divergence loss constrains the output distribution of the edited model to remain close to that of the original frozen model on prompts unrelated to the target edits. The overall training objective is formulated as:
\begin{equation}
\mathcal{L}_{total} = \lambda_{1}\mathcal{L}_{ce} + \lambda_{2}\mathcal{L}_{kl} + \lambda_{3}\mathcal{L}_{gate} + \lambda_{4}\mathcal{L}_{orth},
\end{equation}
where $\lambda_{1}$, $\lambda_{2}$, $\lambda_{3}$, and $\lambda_{4}$ are hyperparameters that balance the contributions of each loss term.

\begin{table*}[t!]
\setlength\tabcolsep{10pt}
\centering
\begin{adjustbox}{width=0.85\textwidth}
\begin{tabular}{clcccc|cccc}
\toprule
\multirow{2}{*}{\textbf{Model}}
& \multirow{2}{*}{\textbf{Method}}
& \multicolumn{4}{c}{\textbf{ZsRE}}
& \multicolumn{4}{c}{\textbf{CounterFact}} \\
\cmidrule(lr){3-6}\cmidrule(lr){7-10}
&
& Eff.$\uparrow$ & Gen.$\uparrow$ & Spe.$\uparrow$ & Avg.$\uparrow$
& Eff.$\uparrow$ & Gen.$\uparrow$ & Spe.$\uparrow$ & Avg.$\uparrow$ \\
\midrule
\multirow{9}{*}{\rotatebox{90}{\textbf{LLaMA-3-8B}}}
& FT
& 26.50 & 25.93 & 15.16 & 22.53 
& \textbf{99.75} & 88.65 & 39.62 & 76.01 \\
& ROME
& 42.58 & 39.84 & 30.70 & 37.71 
& 9.60 & 11.45 & 88.00 & 36.35 \\
& MEMIT
& 87.13 & 84.18 & \textbf{32.11} & 67.81
& 94.55 & 69.55 & 88.31 & 84.14 \\
& PRUNE
& 67.37 & 62.48 & 27.51 & 52.45
& 98.05 & \textbf{95.18} & 74.33 & 89.19 \\
& RECT
& 78.49 & 74.88 & 32.00 & 61.79 
& 75.25 & 50.88 & \textbf{89.24} & 71.79 \\
& NSE
& 45.69 & 44.95 & 31.47 & 40.70 
& 85.70 & 53.95 & 88.35 & 76.00 \\
& AlphaEdit
& 87.37 & 83.93 & 31.95 & 67.75 
& 98.85 & 92.90 & 67.28 & 86.34\\
\cmidrule(lr){2-10}
& ReFT
& 47.65 & 46.48 & 22.82 & 38.98
& 83.50 & 52.32 & 40.26 & 58.69 \\
& ORE (Ours)
& \textbf{94.20} & \textbf{88.98} & 29.90 & \textbf{71.03}
& 98.70 & 92.21 & 83.10 & \textbf{91.34} \\
\midrule
\multirow{9}{*}{\rotatebox{90}{\textbf{Qwen2.5-7B}}}
& FT
& 36.96 & 35.87 & 31.65 & 34.83
& 99.75 & 72.32 & 40.22 & 70.76 \\
& ROME
& 36.52 & 35.42 & 38.34 & 36.76
& 14.00 & 16.75 & \textbf{86.04} & 38.93 \\
& MEMIT
& 95.53 & 90.96 & \textbf{41.72} & 76.07
& 99.50 & 92.45 & 83.61 & 91.85 \\
& PRUNE
& 69.92 & 65.54 & 27.28 & 54.25
& 99.55 & \textbf{98.15} & 72.67 & 90.12 \\
& RECT
& 91.25 & 83.86 & 39.86 & 71.66
& 98.10 & 84.92 & 84.41 & 89.14 \\
& NSE
& 49.65 & 48.78 & 40.81 & 46.41
& 57.45 & 51.35 & 85.18 & 64.66 \\
& AlphaEdit
& 96.49 & 91.47 & 39.04 & 75.67
& \textbf{99.80} & 95.80 & 82.88 & 92.83 \\
\cmidrule(lr){2-10}
& ReFT
& 49.81 & 47.25 & 37.08 & 44.71 
& 77.25 & 48.95 & 48.40 & 58.20 \\
& ORE (Ours)
& \textbf{99.85} & \textbf{94.21} & 35.37 & \textbf{76.48}
& 99.18 & 95.50 & 84.73 & \textbf{93.14} \\

\bottomrule
\end{tabular}
\end{adjustbox}
\caption{Comparison of 2000 edits on ZsRE and CounterFact.}
\label{tab:2000 edits}
\vspace{-8pt}
\end{table*}
\section{Experiments}

\subsection{Experiment Settings}

\noindent \textbf{Datasets.} 
We evaluate ORE on three widely used knowledge editing benchmarks: ZsRE \cite{zsre}, CounterFact \cite{ROME}, and Bi-ZsRE \cite{wang2024cross}. ZsRE is a fact-based question-answering dataset commonly used to assess precise factual updates.
CounterFact is a large-scale and challenging benchmark for counterfactual knowledge editing, featuring diverse relations, entities, paraphrased prompts, and semantically related but factually independent neighborhood samples. Bi-ZsRE is a cross-lingual extension of ZsRE with parallel Chinese–English question-answer pairs, which we use to evaluate ORE in cross-lingual editing scenarios.

\noindent \textbf{Implementation Details and Metrics.}  All experiments are conducted on LLaMA-3-8B and Qwen-2.5-7B models. Please refer to the \textbf{Appendix \ref{appendix:details}} for hyperparameters and more implementation details.
Following prior works \cite{MEMIT, AlphaEdit}, we evaluate models using three standard metrics: \textbf{Efficacy}, \textbf{Generality}, and \textbf{Specificity}. Efficacy measures whether the edited fact is correctly produced for the original prompt; Generality evaluates the robustness of the edited knowledge under paraphrased prompts; and Specificity assesses whether the editing operation avoids affecting factually unrelated knowledge.
\begin{table*}[htbp]
\setlength\tabcolsep{10pt}
\centering
\begin{adjustbox}{width=0.85\textwidth}
\begin{tabular}{clcccc|cccc}
\toprule
\multirow{2}{*}{\textbf{Model}}
& \multirow{2}{*}{\textbf{Method}}
& \multicolumn{4}{c}{\textbf{ZsRE}}
& \multicolumn{4}{c}{\textbf{CounterFact}} \\
\cmidrule(lr){3-6}\cmidrule(lr){7-10}
&
& Eff.$\uparrow$ & Gen.$\uparrow$ & Spe.$\uparrow$ & Avg.$\uparrow$
& Eff.$\uparrow$ & Gen.$\uparrow$ & Spe.$\uparrow$ & Avg.$\uparrow$ \\
\midrule
\multirow{9}{*}{\rotatebox{90}{\textbf{LLaMA-3-8B}}}
& FT
& 37.20 & 36.52 & \textbf{39.94} & 37.89
& \textbf{98.95} & 91.94 & 38.73 & 76.54 \\
& ROME
& 41.45 & 39.35 & 30.64 & 37.15 
& 8.32 & 10.47 & 88.25 & 35.68 \\
& MEMIT
& 87.46 & 84.12 & 31.67 & 67.75
& 96.4 & 76.77 & 86.57 & 86.58 \\
& PRUNE
& 41.73 & 39.50 & 20.94 & 34.06
& 95.29 & 87.75 & 67.48 & 83.51 \\
& RECT
& 68.08 & 64.50 & 30.91 & 54.50 
& 67.18 & 45.69 & \textbf{88.91} & 67.26 \\
& NSE
& 45.21 & 44.47 & 30.58 & 40.09
& 85.22 & 52.33 & 87.49 & 75.01 \\
& AlphaEdit
& 86.35 & 82.73 & 31.10 & 66.73
& 94.07 & 75.15 & 84.69 & 84.64\\
\cmidrule(lr){2-10}
& ReFT
& 44.57 & 43.38 & 21.28 & 36.41
& 80.86 & 55.37 & 40.83 & 59.02 \\
& ORE (Ours)
& \textbf{93.24} & \textbf{88.22} & 29.35 & \textbf{70.27}
& 97.78 & \textbf{96.81} & 76.63 & \textbf{90.41} \\
\midrule
\multirow{9}{*}{\rotatebox{90}{\textbf{Qwen2.5-7B}}}
& FT
& 33.30 & 31.26 & 26.72 & 30.43
& \textbf{99.98} & 75.60 & 38.83 & 71.47 \\
& ROME
& 35.41 & 34.44 & 38.29 & 36.05 
& 12.72 & 15.31 & \textbf{85.98} & 38.00 \\
& MEMIT
& 94.52 & 90.29 & 38.36 & 74.39
& 99.42 & 92.39 & 81.20 & 91.00 \\
& PRUNE
& 69.10 & 65.92 & \textbf{42.00} & 59.01
& 98.36 & \textbf{96.87} & 64.57 & 86.60 \\
& RECT
& 91.01 & 83.99 & 40.71 & 71.90
& 97.72 & 80.04 & 83.33 & 87.03 \\
& NSE
& 49.15 & 48.32 & 40.73 & 46.07 
& 56.72 & 50.14 & 84.01 & 63.62 \\
& AlphaEdit
& 95.23 & 87.75 & 38.97 & 73.98 
& 99.76 & 94.73 & 79.16 & 91.22 \\
\cmidrule(lr){2-10}
& ReFT
& 49.95 & 47.51 & 35.30 & 44.25
& 77.30 & 51.99 & 44.91 & 58.07 \\
& ORE (Ours) 
& \textbf{99.83} & \textbf{93.85} & 30.64 & \textbf{74.77}
& 99.10 & 93.56 & 83.64 & \textbf{92.10} \\
\bottomrule
\end{tabular}
\end{adjustbox}
\caption{Comparison of 5000 edits on ZsRE and CounterFact.}
\label{tab:5000 edits}
\end{table*}

\noindent \textbf{Baselines.} 
We compare ORE with the following baselines. \textbf{FT} \cite{FT} directly fine-tunes selected model parameters on edit samples using cross-entropy loss. \textbf{ROME} \cite{ROME} identifies knowledge-bearing neurons via causal tracing and injects rank-one updates into MLP layers, while \textbf{MEMIT} \cite{MEMIT} extends ROME to support batch editing by distributing update residuals across multiple layers. \textbf{PRUNE} \cite{prune} constrains parameter perturbations by limiting the condition number of the edit matrix to reduce interference during sequential edits, and \textbf{RECT} \cite{rect} improves robustness through consistency regularization to prevent overfitting. \textbf{AlphaEdit} \cite{AlphaEdit} applies null-space projection to inject updates while preserving existing knowledge. 
\textbf{NSE} \cite{NSE} edits models by selectively updating activation-critical neurons based on target hidden states computed from frozen parameters.
\textbf{ReFT} \cite{ReFT} learns lightweight intervention functions that manipulate hidden representations in low-rank subspaces of a frozen model at inference time.

\subsection{Performance Analysis}
Tables~\ref{tab:2000 edits} and~\ref{tab:5000 edits} report comparisons between ORE and representative knowledge editing methods on ZsRE and CounterFact. From the results, we observe that:
(1) Across datasets and model backbones, ORE consistently achieves the competitive performance in terms of editing success and generalization, demonstrating its effectiveness in large-scale batch editing settings. This advantage mainly stems from alleviating general semantic entanglement, which encourages editing directions to focus on semantic dimensions directly relevant to the target facts. ORE is comparatively less dominant in specificity, as its edits are regulated by soft constraints imposed in the representation space rather than hard parameter-level isolation, making it inherently weaker than parameter-editing methods in preserving unrelated knowledge.
(2) Compared with conventional representation fine-tuning, ORE exhibits improved stability and overall performance, validating the effectiveness of subspace orthogonalization combined with the non-linear bottleneck and dynamic gating design.
For completeness, we also report experimental results on sequential knowledge editing in \textbf{Appendix \ref{appendix:seq_edit}}.

\subsection{Analysis of Anti-Interference Capability}

\begin{figure}[t!]
  \centering
  \includegraphics[width=0.4\textwidth]{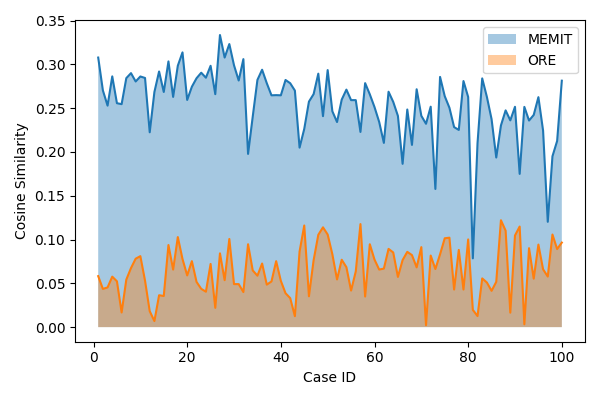}
  \caption{Cosine similarity between edit representations and the general semantic subspace across sequential edits. MEMIT (blue) consistently shows higher similarity, indicating stronger entanglement with shared semantics, while ORE (orange) maintains substantially lower similarity throughout the editing process.}
  \label{fig:memit_vs_ore}
\end{figure}

To validate ORE’s ability to eliminate common semantic entanglement from a geometric perspective, we conducted experiments on the shared structure dataset "The capital of [Country] is [City]" constructed in Section \ref{sec:observation}. Beyond standard Efficacy, we measured the cosine similarity between each method’s edit vectors and the common semantic subspace $S_{gs}$, where higher similarity indicates greater interference from shared semantics.
As shown in Figure~\ref{fig:memit_vs_ore}, MEMIT’s post-edit vectors remain highly aligned with $S_{gs}$, suggesting that its updates fail to avoid shared semantic regions. In contrast, ORE, via representation orthogonal projection, achieves 92\% Efficacy while reducing cosine similarity to only one-sixth of MEMIT’s, demonstrating its effectiveness in mitigating representation entanglement and enhancing anti-interference capability in batch editing.

\subsection{Ablation Study}

\begin{table}[t!]
\setlength\tabcolsep{10pt}
\centering
\begin{adjustbox}{width=0.4\textwidth}
\begin{tabular}{lcccc}
\toprule
\textbf{Method}
& Eff.$\uparrow$ & Gen.$\uparrow$ & Spe.$\uparrow$ & Avg.$\uparrow$ \\
\midrule
ORE
& 95.54 & 92.18 & 29.03 & 72.25 \\
-$\mathcal{L}_{orth}$
& 93.34 & 88.77 & 29.18 & 70.43 \\
-SiLU
& 93.00 & 90.52 & 27.46 & 70.33 \\
-Gate
& 95.13 & 85.78 & 28.64 & 69.85 \\
\bottomrule
\end{tabular}
\end{adjustbox}
\caption{Ablation study of ORE with 1,000 edits on ZsRE.}
\label{tab:ablation}
\end{table}

\begin{figure}[t!]
  \centering
  \includegraphics[width=0.4\textwidth]{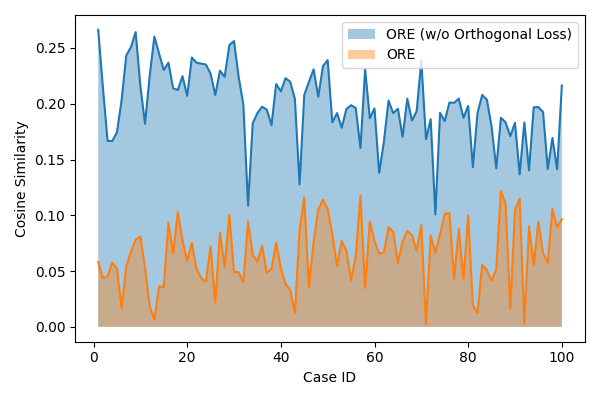}
  \caption{Ablation study of Representation Subspace Orthogonalization Loss.}
  \label{fig:ablation_orth}
\end{figure}

We conduct an ablation study on ZsRE under a batch editing setting with 1,000 samples to evaluate the contributions of ORE’s key components. As shown in Table~\ref{tab:ablation}, removing the representation orthogonalization loss (-$\mathcal{L}_{orth}$) decreases both efficacy and generality, indicating that orthogonalization is crucial for decoupling shared semantic patterns and mitigating cumulative interference. Eliminating the non-linear activation in the gated representation head (-SiLU) leads to a drop in overall performance. Disabling the dynamic gating mechanism (-Gate) reduces generality and specificity, highlighting the importance of adaptive gating for accurately localizing knowledge-carrying positions while avoiding unnecessary perturbations in irrelevant regions.
Figure~\ref{fig:ablation_orth} further shows the cosine similarity between edit vectors and the general semantic subspace after removing the representation orthogonalization, illustrating that the orthogonal constraint effectively decouples the edited facts from shared semantic patterns.

\subsection{Cross-Lingual Editing Scenarios}

Furthermore, we evaluated the performance of ORE in cross-lingual editing scenarios. Table \ref{tab:1600 edits} shows the comparative results with an editing batch size of 1600. The experimental results demonstrate that ORE outperforms existing methods on the Bi-ZsRE benchmark, achieving an average score of 65.36\%. Notably, this represents an improvement of 4.40 percentage points compared to the SOTA method MEMIT (60.96\%). This indicates that ORE maintains high knowledge editing accuracy and generalization capabilities even in challenging mixed-language scenarios, successfully mitigating the interference typically caused by general semantic spaces across languages.

\begin{table}[t!]
\centering
\begin{adjustbox}{width=0.9\linewidth}
\begin{tabular}{clcccc}
\toprule
\multirow{2}{*}{\textbf{Model}}
& \multirow{2}{*}{\textbf{Method}}
& \multicolumn{4}{c}{\textbf{Bi-ZsRE}} \\
\cmidrule(lr){3-6}
&
& Eff.$\uparrow$ & Gen.$\uparrow$ & Spe.$\uparrow$ & Avg.$\uparrow$ \\
\midrule
\multirow{9}{*}{\rotatebox{90}{\textbf{LLaMA-3-8B}}}
& FT 
& 48.52 & 45.06 & 22.47 & 38.68 \\
& ROME
& 42.08 & 40.80 & 30.18 & 37.69 \\
& MEMIT 
& 76.93 & 74.55 & \textbf{31.40} & 60.96 \\
& PRUNE
& 53.09 & 50.59 & 23.12 & 42.27 \\
& RECT
& 67.78 & 65.06 & 31.39 & 54.74 \\
& NSE
& 49.46 & 48.51 & 30.76 & 42.91 \\
& AlphaEdit 
& 76.14 & 73.30 & 31.23 & 60.22\\
\cmidrule(lr){2-6}
& ReFT
& 53.67 & 52.29 & 24.49 & 43.48 \\
& ORE (Ours)
& \textbf{85.10} & \textbf{81.84} & 29.14 & \textbf{65.36} \\
\midrule
\multirow{9}{*}{\rotatebox{90}{\textbf{Qwen2.5-7B}}}
& FT
& 47.25 & 42.39 & 33.32 & 40.99 \\
& ROME
& 37.02 & 36.66 & 39.72 & 37.80 \\
& MEMIT
& 79.45 & 76.52 & 40.77 & 65.58 \\
& PRUNE
& 67.39 & 64.91 & \textbf{43.07} & 58.46 \\
& RECT
& 76.36 & 72.23 & 41.00 & 63.20 \\
& NSE
& 52.54 & 51.27 & 38.79 & 47.53 \\
& AlphaEdit
& 82.28 & 78.49 & 39.81 & 66.86 \\
\cmidrule(lr){2-6}
& ReFT
& 52.89 & 50.37 & 35.27 & 46.18 \\
& ORE (Ours) 
& \textbf{86.49} & \textbf{83.45} & 31.04 & \textbf{66.99} \\
\bottomrule
\end{tabular}
\end{adjustbox}
\caption{Comparison of 1600 edits on Bi-ZsRE.}
\label{tab:1600 edits}
\end{table}

\section{Related Work}

Existing knowledge editing methods can be broadly categorized into two groups: parameter-modifying and parameter-preserving methods.

Parameter-modifying methods inject new knowledge by locating and updating specific model weights. ROME \cite{ROME} introduced a "Locate-Then-Edit" paradigm with rank-one updates to MLP layers for single facts, and MEMIT \cite{MEMIT} extended this to batch editing by distributing update residuals across layers. To mitigate interference and preserve general capabilities, RECT \cite{rect} adds regularization, PRUNE \cite{prune} controls the condition number of edit weights, AlphaEdit \cite{AlphaEdit} projects updates onto the null space of retained knowledge, LangEdit \cite{LangEdit} and KDE \cite{KDE} project updates onto dynamic or orthogonal subspaces to reduce cross-lingual or lifelong interference, and AdaEdit \cite{AdaEdit} addresses sequential decline via disentangled FFN representations and SVD-based sparsification. While KDE and LangEdit also introduce orthogonality constraints, they enforce orthogonality on parameter updates, whereas ORE operates directly in the representation space to decouple semantic entanglement.

Parameter-preserving methods keep the backbone frozen and introduce external modules or representation interventions. T-Patcher \cite{huang2023transformer} adds task-specific parameters, while SERAC \cite{serac} and GRACE \cite{GRACE} use memory-based modules for non-intrusive updates. Recently, ReFT \cite{ReFT} demonstrates that low-rank interventions on hidden states at inference time are sufficient to steer model behavior, and BaFT \cite{BaFT} proposed an input-based basis vector weighting mechanism, achieving a better trade-off between editing and locality through non-linear, fine-grained control of the representation subspace. 
In comparison, the proposed ORE follows the ReFT paradigm and improves upon it to adapt to knowledge editing scenarios. Addressing the issue of general semantic entanglement in batch editing, we introduce explicit geometric orthogonal constraints on top of representation intervention, achieving great performance in batch and cross-lingual scenarios.
\section{Conclusion}
In this work, we identify that general semantic structure entanglement negatively impacts batch knowledge editing in LLMs. To address this issue, we propose Orthogonal Representation Editing (ORE), a parameter-preserving knowledge editing framework that operates directly on representation vectors by constructing a general semantic subspace and enforcing orthogonal editing directions to decouple shared semantics, together with a gated non-linear representation tuning mechanism for precise and localized representation intervention. Extensive experiments demonstrate that ORE consistently outperforms existing methods in both editing accuracy and generalization, including challenging cross-lingual editing scenarios.

\section*{Limitations}

Despite ORE demonstrating superior performance in batch editing and cross-lingual scenarios, several limitations remain that merit further exploration in future work: 
(1) Our current experiments are primarily conducted on models with 7B to 8B parameters. Larger-scale models typically possess higher-dimensional representation spaces and more complex entanglement patterns. Consequently, the geometric constraint mechanism proposed in ORE requires further experimental verification on these larger-scale architectures. 
(2) Current experiments are mainly concentrated on standard and restricted editing settings. Future work should extend the evaluation scope to a broader range of downstream application scenarios to verify the feasibility and stability of the method in complex and dynamic real-world deployments.
(3) Our current work primarily focuses on improving the accuracy of the model in answering questions, with less emphasis on capabilities such as logical reasoning following the editing process. Future research should further investigate the impact of knowledge editing on the model's comprehensive reasoning abilities.
\section{Acknowledgments}
The work is supported by the National Natural Science Foundation of China (Grant 62406223) 
\bibliography{main}

@inproceedings{KDE,
    title = "Knowledge Decoupling via Orthogonal Projection for Lifelong Editing of Large Language Models",
    author = "Xu, Haoyu  and
      Lan, Pengxiang  and
      Yang, Enneng  and
      Guo, Guibing  and
      Zhao, Jianzhe  and
      Jiang, Linying  and
      Wang, Xingwei",
    editor = "Che, Wanxiang  and
      Nabende, Joyce  and
      Shutova, Ekaterina  and
      Pilehvar, Mohammad Taher",
    booktitle = "Proceedings of the 63rd Annual Meeting of the Association for Computational Linguistics (Volume 1: Long Papers)",
    month = jul,
    year = "2025",
    address = "Vienna, Austria",
    publisher = "Association for Computational Linguistics",
    url = "https://aclanthology.org/2025.acl-long.646/",
    doi = "10.18653/v1/2025.acl-long.646",
    pages = "13194--13213",
    ISBN = "979-8-89176-251-0"
}

@inproceedings{
AlphaEdit,
title={AlphaEdit: Null-Space Constrained Model Editing for Language Models},
author={Junfeng Fang and Houcheng Jiang and Kun Wang and Yunshan Ma and Jie Shi and Xiang Wang and Xiangnan He and Tat-Seng Chua},
booktitle={The Thirteenth International Conference on Learning Representations},
year={2025},
url={https://openreview.net/forum?id=HvSytvg3Jh}
}

@inproceedings{LangEdit,
    title = "Mitigating Negative Interference in Multilingual Knowledge Editing through Null-Space Constraints",
    author = "Sun, Wei  and
      Qu, Tingyu  and
      Li, Mingxiao  and
      Davis, Jesse  and
      Moens, Marie-Francine",
    editor = "Che, Wanxiang  and
      Nabende, Joyce  and
      Shutova, Ekaterina  and
      Pilehvar, Mohammad Taher",
    booktitle = "Findings of the Association for Computational Linguistics: ACL 2025",
    month = jul,
    year = "2025",
    address = "Vienna, Austria",
    publisher = "Association for Computational Linguistics",
    url = "https://aclanthology.org/2025.findings-acl.460/",
    doi = "10.18653/v1/2025.findings-acl.460",
    pages = "8796--8810",
    ISBN = "979-8-89176-256-5"
}

@inproceedings{AdaEdit,
    title = "{A}da{E}dit: Advancing Continuous Knowledge Editing For Large Language Models",
    author = "Li, Qi  and
      Chu, Xiaowen",
    editor = "Che, Wanxiang  and
      Nabende, Joyce  and
      Shutova, Ekaterina  and
      Pilehvar, Mohammad Taher",
    booktitle = "Proceedings of the 63rd Annual Meeting of the Association for Computational Linguistics (Volume 1: Long Papers)",
    month = jul,
    year = "2025",
    address = "Vienna, Austria",
    publisher = "Association for Computational Linguistics",
    url = "https://aclanthology.org/2025.acl-long.208/",
    doi = "10.18653/v1/2025.acl-long.208",
    pages = "4127--4149",
    ISBN = "979-8-89176-251-0"
}

@inproceedings{
ReFT,
title={Re{FT}: Representation Finetuning for Language Models},
author={Zhengxuan Wu and Aryaman Arora and Zheng Wang and Atticus Geiger and Dan Jurafsky and Christopher D Manning and Christopher Potts},
booktitle={The Thirty-eighth Annual Conference on Neural Information Processing Systems},
year={2024},
url={https://openreview.net/forum?id=fykjplMc0V}
}

@inproceedings{NSE,
    title = "Neuron-Level Sequential Editing for Large Language Models",
    author = "Jiang, Houcheng  and
      Fang, Junfeng  and
      Zhang, Tianyu  and
      Bi, Baolong  and
      Zhang, An  and
      Wang, Ruipeng  and
      Liang, Tao  and
      Wang, Xiang",
    editor = "Che, Wanxiang  and
      Nabende, Joyce  and
      Shutova, Ekaterina  and
      Pilehvar, Mohammad Taher",
    booktitle = "Proceedings of the 63rd Annual Meeting of the Association for Computational Linguistics (Volume 1: Long Papers)",
    month = jul,
    year = "2025",
    address = "Vienna, Austria",
    publisher = "Association for Computational Linguistics",
    url = "https://aclanthology.org/2025.acl-long.815/",
    doi = "10.18653/v1/2025.acl-long.815",
    pages = "16678--16702",
    ISBN = "979-8-89176-251-0"
}

@inproceedings{
MEMIT,
title={Mass-Editing Memory in a Transformer},
author={Kevin Meng and Arnab Sen Sharma and Alex J Andonian and Yonatan Belinkov and David Bau},
booktitle={The Eleventh International Conference on Learning Representations },
year={2023},
url={https://openreview.net/forum?id=MkbcAHIYgyS}
}

@inproceedings{
GRACE,
title={Aging with {GRACE}: Lifelong Model Editing with Discrete Key-Value Adaptors},
author={Thomas Hartvigsen and Swami Sankaranarayanan and Hamid Palangi and Yoon Kim and Marzyeh Ghassemi},
booktitle={Thirty-seventh Conference on Neural Information Processing Systems},
year={2023},
url={https://openreview.net/forum?id=Oc1SIKxwdV}
}

@inproceedings{
BaFT,
title={Unlocking Efficient, Scalable, and Continual Knowledge Editing with Basis-Level Representation Fine-Tuning},
author={Tianci Liu and Ruirui Li and Yunzhe Qi and Hui Liu and Xianfeng Tang and Tianqi Zheng and Qingyu Yin and Monica Xiao Cheng and Jun Huan and Haoyu Wang and Jing Gao},
booktitle={The Thirteenth International Conference on Learning Representations},
year={2025},
url={https://openreview.net/forum?id=PITFO1ddeh}
}

@inproceedings{
ROME,
title={Locating and Editing Factual Associations in {GPT}},
author={Kevin Meng and David Bau and Alex J Andonian and Yonatan Belinkov},
booktitle={Advances in Neural Information Processing Systems},
editor={Alice H. Oh and Alekh Agarwal and Danielle Belgrave and Kyunghyun Cho},
year={2022},
url={https://openreview.net/forum?id=-h6WAS6eE4}
}

@article{
  backtrack,
  title={Experience replay for continual learning},
  author={Rolnick, David and Ahuja, Arun and Schwarz, Jonathan and Lillicrap, Timothy and Wayne, Gregory},
  journal={Advances in neural information processing systems},
  volume={32},
  year={2019}
}

@article{mann2020language,
  title={Language models are few-shot learners},
  author={Mann, Ben and Ryder, Nick and Subbiah, Melanie and Kaplan, J and Dhariwal, P and Neelakantan, A and Shyam, P and Sastry, G and Askell, A and Agarwal, S and others},
  journal={arXiv preprint arXiv:2005.14165},
  volume={1},
  number={3},
  pages={3},
  year={2020}
}

@article{brown2020language,
  title={Language models are few-shot learners},
  author={Brown, Tom and Mann, Benjamin and Ryder, Nick and Subbiah, Melanie and Kaplan, Jared D and Dhariwal, Prafulla and Neelakantan, Arvind and Shyam, Pranav and Sastry, Girish and Askell, Amanda and others},
  journal={Advances in neural information processing systems},
  volume={33},
  pages={1877--1901},
  year={2020}
}

@article{wang2024knowledge,
  title={Knowledge editing for large language models: A survey},
  author={Wang, Song and Zhu, Yaochen and Liu, Haochen and Zheng, Zaiyi and Chen, Chen and Li, Jundong},
  journal={ACM Computing Surveys},
  volume={57},
  number={3},
  pages={1--37},
  year={2024},
  publisher={ACM New York, NY}
}

@article{yao2023editing,
  title={Editing large language models: Problems, methods, and opportunities},
  author={Yao, Yunzhi and Wang, Peng and Tian, Bozhong and Cheng, Siyuan and Li, Zhoubo and Deng, Shumin and Chen, Huajun and Zhang, Ningyu},
  journal={arXiv preprint arXiv:2305.13172},
  year={2023}
}

@article{gupta2024model,
  title={Model editing at scale leads to gradual and catastrophic forgetting},
  author={Gupta, Akshat and Rao, Anurag and Anumanchipalli, Gopala},
  journal={arXiv preprint arXiv:2401.07453},
  year={2024}
}

@article{REMEDI,
  title={Inspecting and editing knowledge representations in language models},
  author={Hernandez, Evan and Li, Belinda Z and Andreas, Jacob},
  journal={arXiv preprint arXiv:2304.00740},
  year={2023}
}

@article{huang2023transformer,
  title={Transformer-patcher: One mistake worth one neuron},
  author={Huang, Zeyu and Shen, Yikang and Zhang, Xiaofeng and Zhou, Jie and Rong, Wenge and Xiong, Zhang},
  journal={arXiv preprint arXiv:2301.09785},
  year={2023}
}

@article{jiang2025anyedit,
  title={Anyedit: Edit any knowledge encoded in language models},
  author={Jiang, Houcheng and Fang, Junfeng and Zhang, Ningyu and Ma, Guojun and Wan, Mingyang and Wang, Xiang and He, Xiangnan and Chua, Tat-seng},
  journal={arXiv preprint arXiv:2502.05628},
  year={2025}
}

@inproceedings{beniwal2024cross,
  title={Cross-lingual editing in multilingual language models},
  author={Beniwal, Himanshu and Kowsik, D and Singh, Mayank},
  booktitle={Findings of the Association for Computational Linguistics: EACL 2024},
  pages={2078--2128},
  year={2024}
}

@inproceedings{wang2024cross,
  title={Cross-lingual knowledge editing in large language models},
  author={Wang, Jiaan and Liang, Yunlong and Sun, Zengkui and Cao, Yuxuan and Xu, Jiarong and Meng, Fandong},
  booktitle={Proceedings of the 62nd Annual Meeting of the Association for Computational Linguistics (Volume 1: Long Papers)},
  pages={11676--11686},
  year={2024}
}

@article{cai2024edit,
  title={O-edit: Orthogonal subspace editing for language model sequential editing},
  author={Cai, Yuchen and Cao, Ding},
  journal={arXiv preprint arXiv:2410.11469},
  year={2024}
}

@article{wang2025decoupling,
  title={Decoupling Reasoning and Knowledge Injection for In-Context Knowledge Editing},
  author={Wang, Changyue and Su, Weihang and Ai, Qingyao and Zhou, Yujia and Liu, Yiqun},
  journal={arXiv preprint arXiv:2506.00536},
  year={2025}
}

@article{zsre,
  title={Zero-shot relation extraction via reading comprehension},
  author={Levy, Omer and Seo, Minjoon and Choi, Eunsol and Zettlemoyer, Luke},
  journal={arXiv preprint arXiv:1706.04115},
  year={2017}
}

@inproceedings{
prune,
title={Perturbation-Restrained Sequential Model Editing},
author={Jun-Yu Ma and Hong Wang and Hao-Xiang Xu and Zhen-Hua Ling and Jia-Chen Gu},
booktitle={The Thirteenth International Conference on Learning Representations},
year={2025},
url={https://openreview.net/forum?id=bfI8cp8qmk}
}

@inproceedings{rect,
    title = "Model Editing Harms General Abilities of Large Language Models: Regularization to the Rescue",
    author = "Gu, Jia-Chen  and
      Xu, Hao-Xiang  and
      Ma, Jun-Yu  and
      Lu, Pan  and
      Ling, Zhen-Hua  and
      Chang, Kai-Wei  and
      Peng, Nanyun",
    editor = "Al-Onaizan, Yaser  and
      Bansal, Mohit  and
      Chen, Yun-Nung",
    booktitle = "Proceedings of the 2024 Conference on Empirical Methods in Natural Language Processing",
    month = nov,
    year = "2024",
    address = "Miami, Florida, USA",
    publisher = "Association for Computational Linguistics",
    url = "https://aclanthology.org/2024.emnlp-main.934/",
    doi = "10.18653/v1/2024.emnlp-main.934",
    pages = "16801--16819"
}

@inproceedings{serac,
  title={Memory-based model editing at scale},
  author={Mitchell, Eric and Lin, Charles and Bosselut, Antoine and Manning, Christopher D and Finn, Chelsea},
  booktitle={International Conference on Machine Learning},
  pages={15817--15831},
  year={2022},
  organization={PMLR}
}

@article{FT,
  title={Modifying memories in transformer models},
  author={Zhu, Chen and Rawat, Ankit Singh and Zaheer, Manzil and Bhojanapalli, Srinadh and Li, Daliang and Yu, Felix and Kumar, Sanjiv},
  journal={arXiv preprint arXiv:2012.00363},
  year={2020}
}

@article{abdi2010principal,
  title={Principal component analysis},
  author={Abdi, Herv{\'e} and Williams, Lynne J},
  journal={Wiley interdisciplinary reviews: computational statistics},
  volume={2},
  number={4},
  pages={433--459},
  year={2010},
  publisher={Wiley Online Library}
}

@article{hu2022lora,
  title={Lora: Low-rank adaptation of large language models.},
  author={Hu, Edward J and Shen, Yelong and Wallis, Phillip and Allen-Zhu, Zeyuan and Li, Yuanzhi and Wang, Shean and Wang, Lu and Chen, Weizhu and others},
  journal={ICLR},
  volume={1},
  number={2},
  pages={3},
  year={2022}
}

@inproceedings{SAKE,
    title = "{SAKE}: Steering Activations for Knowledge Editing",
    author = "Scialanga, Marco  and
      Laugel, Thibault  and
      Grari, Vincent  and
      Detyniecki, Marcin",
    editor = "Che, Wanxiang  and
      Nabende, Joyce  and
      Shutova, Ekaterina  and
      Pilehvar, Mohammad Taher",
    booktitle = "Proceedings of the 63rd Annual Meeting of the Association for Computational Linguistics (Volume 1: Long Papers)",
    month = jul,
    year = "2025",
    address = "Vienna, Austria",
    publisher = "Association for Computational Linguistics",
    url = "https://aclanthology.org/2025.acl-long.777/",
    doi = "10.18653/v1/2025.acl-long.777",
    pages = "15966--15978",
    ISBN = "979-8-89176-251-0"
}

@inproceedings{deck,
    title = "Decoding by Contrasting Knowledge: Enhancing Large Language Model Confidence on Edited Facts",
    author = "Bi, Baolong  and
      Liu, Shenghua  and
      Mei, Lingrui  and
      Wang, Yiwei  and
      Fang, Junfeng  and
      Ji, Pengliang  and
      Cheng, Xueqi",
    editor = "Che, Wanxiang  and
      Nabende, Joyce  and
      Shutova, Ekaterina  and
      Pilehvar, Mohammad Taher",
    booktitle = "Proceedings of the 63rd Annual Meeting of the Association for Computational Linguistics (Volume 1: Long Papers)",
    month = jul,
    year = "2025",
    address = "Vienna, Austria",
    publisher = "Association for Computational Linguistics",
    url = "https://aclanthology.org/2025.acl-long.841/",
    doi = "10.18653/v1/2025.acl-long.841",
    pages = "17198--17208",
    ISBN = "979-8-89176-251-0"
}

\appendix

\section{Datasets}
To comprehensively evaluate the performance of ORE, we conducted experiments on three widely used knowledge editing benchmark datasets:

\noindent\textbf{ZsRE:} A standard fact-based dataset in question-answering format. Each sample contains a natural language question and its corresponding target answer \cite{zsre}.

\noindent\textbf{CounterFact:} A large-scale and highly challenging benchmark for counterfactual knowledge editing. This dataset covers diverse relation types and entities, and equips each editing target with semantically equivalent paraphrase prompts as well as semantically related but factually independent neighborhood samples \cite{ROME}.

\noindent\textbf{Bi-ZsRE:} A cross-lingual extension of ZsRE, containing parallel Chinese-English question-answer pairs. Given that ORE aims to address the problem of general concept entanglement, we introduce this dataset to specifically evaluate the model's performance in cross-lingual scenarios, examining whether it can effectively strip away linguistic noise and achieve precise cross-lingual knowledge synchronization within the semantic space shared by multilingual models \cite{wang2024cross}.

We follow the experimental settings of \cite{AlphaEdit} for ZsRE and CounterFact, and those of \cite{LangEdit} for Bi-ZsRE.

\section{Metrics}
Following the standards of previous knowledge editing works \cite{ROME, MEMIT, AlphaEdit, LangEdit}, Let the given edit sample be denoted as $(s_i, r_i, o^*_i)$, the prompt as $(s_i, r_i)$, the paraphrase prompts as $N(s_i, r_i)$, and the neighborhood prompts as $O(s_i, r_i)$. The metrics are defined as follows:

\noindent\textbf{Efficacy:} Measures whether the target knowledge has been successfully injected into the model. 

\begin{equation}
    \mathbb{E}_{i} \left\{ o_{i} = \arg \max_{o} \mathbb{P}_{f} (o \mid (s_{i}, r_{i})) \right\}
\end{equation}

\noindent\textbf{Generality:} Measures the robustness of the edited knowledge to semantic variations.

\begin{equation}
    \mathbb{E}_{i} \left\{ o_{i} = \arg \max_{o} \mathbb{P}_{f} (o \mid N((s_{i}, r_{i}))) \right\}
\end{equation}

\noindent\textbf{Specificity:} Measures the locality of the editing operation, evaluating whether the model avoids corrupting irrelevant knowledge.

\begin{equation}
    \mathbb{E}_{i} \left\{ o_{i}^{c} = \arg \max_{o} \mathbb{P}_{f} (o \mid O((s_{i}, r_{i}))) \right\}
\end{equation}

\section{Implementation Details}
\label{appendix:details}
All experiments are conducted on LLaMA-3-8B and Qwen-2.5-7B models. For the LLaMA-3-8B model, we apply interventions at layers [9, 18, 24, 28]; For the Qwen2.5-7B model, we apply interventions at layers [9, 18, 24, 26]. The general semantic subspace is constructed from $2000$ structurally similar but factually unrelated samples. PCA is applied to their representations, and the top $4$ principal components are retained to define the subspace. All experiments were conducted on a single Ascend 910B NPU (64GB).

For all experiments, the loss weights are set to $\lambda_{1}=1.0$, $\lambda_{2}=2.0$, $\lambda_{3}=1.0$, and $\lambda_{4}=3.0$. 
The projection dimension of the non-linear bottleneck is fixed to 128. 
Models are trained for 30 epochs with a batch size of 1 using the AdamW optimizer. 
We adopt a cosine annealing learning rate schedule, with the learning rate decayed from $5\times10^{-4}$ to $2\times10^{-6}$ over the course of training.

For baseline methods, we follow the hyperparameter settings reported in prior work. Specifically, the hyperparameters of FT, ROME, MEMIT, PRUNE, RECT, AlphaEdit, and NSE are adopted from \cite{AlphaEdit}, while ReFT follows \cite{BaFT}.

All experiments are repeated three times, and the reported results are averaged over the three runs.

\begin{figure*}[htp]
  \centering
  \includegraphics[width=1\textwidth]{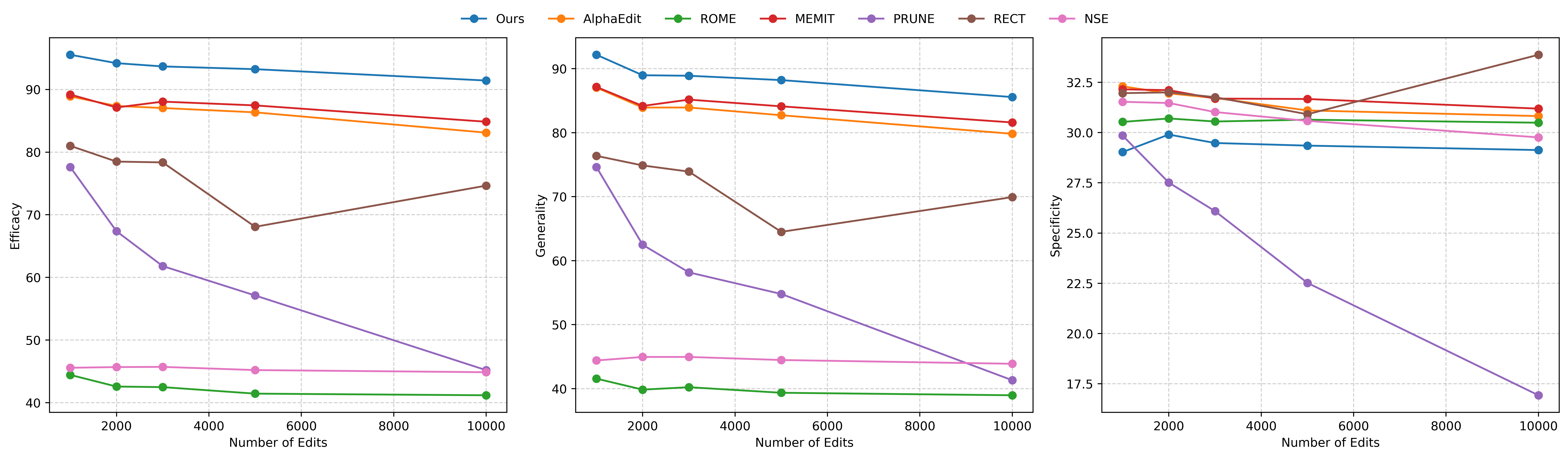}
  \caption{Performance Comparison between ORE and Existing Methods.}
  \label{fig:scale}
\end{figure*}

\section{Baselines}

We compare ORE against the following existing knowledge editing methods:

\noindent\textbf{FT (Fine-Tuning):} The standard fine-tuning approach. We directly update the parameters of specific layers in the model using the cross-entropy loss on the edit samples. FT serves as a fundamental baseline for measuring editing performance \cite{FT}.

\noindent\textbf{ROME (Rank-One Model Editing):} A classic "Locate-Then-Edit" method. ROME employs causal tracing to locate the critical neurons responsible for storing knowledge and uses a rank-one approximation to inject specific key-value pairs into the MLP layers \cite{ROME}.

\noindent\textbf{MEMIT (Mass-Editing Memory in a Transformer):} As an extension of ROME, MEMIT achieves the batch injection of thousands of knowledge entries by distributing the update residuals across the MLP modules of multiple layers \cite{MEMIT}.

\noindent\textbf{PRUNE (Perturbation Restraint on Upper bouNd
for Editing) :} An editing framework constrains the range of parameter perturbation by limiting the condition number of the edit matrix, thereby reducing damage to irrelevant knowledge during sequential editing processes \cite{prune}.

\noindent\textbf{RECT (RElative Change in weighT) :} A method focusing on editing consistency and robustness by introducing an additional regularization term during optimization to prevent weight updates from overfitting to the edit samples \cite{rect}.

\noindent\textbf{AlphaEdit:} The latest SOTA method among parameter-modifying approaches. AlphaEdit proposes a Null-Space Projection mechanism, achieving efficient updates without disrupting the original knowledge \cite{AlphaEdit}.

\noindent\textbf{NSE (Neuron-level Sequential Editing):} NSE utilizes original parameters to calculate target hidden states to prevent model collapse. It also employs an activation-based neuron filtering strategy to update only critical neurons, thereby alleviating catastrophic forgetting \cite{NSE}.

\section{Experimental Results on Sequential Editing}
\label{appendix:seq_edit}

\begin{table}[t]
\centering
\begin{adjustbox}{width=\linewidth}
\begin{tabular}{lcccc}
\toprule
\textbf{Method}
& Eff.$\uparrow$ & Gen.$\uparrow$ & Spe.$\uparrow$ & Avg.$\uparrow$ \\
\midrule

FT
& 30.48 & 30.22 & 15.49 & 25.40 \\

ROME
& 2.01 & 1.80 & 0.69 & 1.50 \\

MEMIT
& 34.62 & 31.28 & 18.49 & 28.13 \\

PRUNE
& 24.77 & 23.87 & 20.69 & 23.11 \\

RECT
& 86.05 & 80.54 & 31.67 & 66.09 \\

AlphaEdit
& 94.47 & \textbf{91.13} & \textbf{32.55} & \textbf{72.72} \\

NSE
& 62.29 & 47.13 & 32.32 & 47.25 \\

\midrule
ReFT
& 19.47 & 18.77 & 13.88 & 17.37 \\

ORE (Ours)
& \textbf{94.71} & 90.86 & 29.24 & 71.60 \\

\bottomrule
\end{tabular}
\end{adjustbox}
\caption{Sequential Editing on the ZsRE benchmark using LLaMA-3-8B.}
\label{tab:zsre_llama3}
\end{table}

Although ORE is primarily designed to address general semantic entanglement in batch knowledge editing, we additionally report results on sequential editing as a complementary evaluation. All the experiment settings are followed \cite{AlphaEdit} and \cite{LangEdit}.

To mitigate catastrophic forgetting under sequential edits, we incorporate an experience replay mechanism following prior continual learning practice \cite{backtrack}.
Specifically, we maintain a replay buffer consisting of previously edited requests.
During training for the current edits, we uniformly sample a small subset from the buffer, where the replay size is 1.
The overall objective at each optimization step is augmented by an additional replay loss computed on the sampled historical edits: $\mathcal{L}' = \mathcal{L} + \mathcal{L}_{replay}$.

\begin{table}[htbp]
\centering
\begin{adjustbox}{width=\linewidth}
\begin{tabular}{lcccc}
\toprule
\textbf{Method}
& Eff.$\uparrow$ & Gen.$\uparrow$ & Spe.$\uparrow$ & Avg.$\uparrow$ \\
\midrule

FT
& 83.33 & 67.79 & 46.63 & 65.92 \\

ROME
& 64.40 & 61.42 & 49.44 & 58.42 \\

MEMIT
& 65.65 & 64.65 & 51.56 & 60.62 \\

PRUNE
& 68.25 & 64.75 & 49.82 & 60.94 \\

RECT
& 66.05 & 63.62 & 61.41 & 63.69 \\

AlphaEdit
& \textbf{98.90} & \textbf{94.22} & 67.88 & 87.00 \\

NSE
& 96.14 & 78.42 & \textbf{87.66} & \textbf{87.41} \\

\midrule
ReFT
& 42.03 & 42.46 & 56.50 & 47.00 \\

ORE (Ours)
& 86.76 & 84.20 & 85.26 & 85.41 \\

\bottomrule
\end{tabular}
\end{adjustbox}
\caption{Sequential Editing on the CounterFact benchmark using LLaMA-3-8B.}
\label{tab:cf_llama3}
\end{table}

Table \ref{tab:zsre_llama3}, \ref{tab:cf_llama3} and \ref{tab:bzsre_llama3} report the performance of different knowledge editing methods under sequential editing settings on ZsRE, CounterFact, and Bi-ZsRE benchmarks using LLaMA3-8B as the backbone model. From the results, we observe that:
(1) ORE achieves competitive overall performance across all three benchmarks and outperforms all the baselines on the Bi-ZsRE dataset, indicating its effectiveness in complex knowledge editing scenarios.
(2) Interestingly, ORE exhibits higher Specificity in sequential editing compared to batch editing. This behavior can be attributed to the dynamic gating mechanism. In sequential settings, each editing step focuses on a smaller batch, making the knowledge-carrying token positions more clearly identifiable. This allows the gating network to activate intervention in a highly selective and sparse manner, while remaining silent on irrelevant tokens and non-target contexts.

\begin{table}[htbp]
\centering
\begin{adjustbox}{width=\linewidth}
\begin{tabular}{lcccc}
\toprule
\textbf{Method}
& Eff.$\uparrow$ & Gen.$\uparrow$ & Spe.$\uparrow$ & Avg.$\uparrow$ \\
\midrule

FT
& 31.41 & 29.97 & 15.29 & 25.56 \\

ROME
& 2.54 & 2.46 & 0.39 & 1.80 \\

MEMIT
& 4.58 & 4.03 & 2.84 & 3.82 \\

PRUNE
& 4.92 & 4.22 & 1.90 & 3.68 \\

RECT
& 41.01 & 38.58 & 20.80 & 33.46 \\

AlphaEdit
& 71.88 & 66.55 & 30.47 & 56.30 \\

LangEdit
& 73.18 & 66.95 & \textbf{31.11} & 57.08 \\

NSE
& 49.43 & 48.06 & 30.58 & 42.69 \\

\midrule
ReFT
& 26.07 & 25.54 & 15.22 & 22.28 \\

ORE (Ours) 
& \textbf{80.53} & \textbf{76.31} & 26.81 & \textbf{61.22} \\

\bottomrule
\end{tabular}
\end{adjustbox}
\caption{Sequential Editing on the Bi-ZsRE benchmark using LLaMA-3-8B.}
\label{tab:bzsre_llama3}
\end{table}

\section{Impact of Batch Editing Scale on Editing Performance}

Figure \ref{fig:scale} illustrates the trends of Efficacy, Generality, and Specificity as the batch editing scale increases for different baselines. As the batch size further expands, the overall performance of most baselines exhibits a clear degradation, reflecting the fact that large-scale editing inevitably introduces general semantic entanglement, which weakens the model’s ability to precisely control target facts. In contrast, ORE demonstrates a notably more stable performance trend and is able to maintain over 90\% Efficacy and Generality even under 10,000 edits. These results validate the effectiveness of ORE in mitigating general semantic entanglement through representation-space orthogonal constraints, enabling more reliable performance scaling in large-scale batch knowledge editing scenarios.

\end{document}